\documentclass[journal,a4paper,twoside]{IEEEtran}

\usepackage{standalone,import,pgfplots,pdfpages,graphicx,float}
\pgfplotsset{width=9cm, height=5.60cm,compat=1.7}
\usepgfplotslibrary{external}
\usepackage[utf8]{inputenc}
\usepackage[counterclockwise]{rotating}
\usepackage{epstopdf}
\usepackage[T1]{fontenc}
\usepackage{lipsum}
\usepackage{epsfig}

\usepackage{cite}       

\usepackage{graphicx}  

\usepackage{epsfig}
\usepackage{psfrag}    

\usepackage{subfigure} 

\usepackage{url}       

\usepackage{stfloats}  
\usepackage{amssymb}
\usepackage{amsmath}   

\usepackage{array}

\usepackage{euscript}
\usepackage{eufrak}
\usepackage{bm}
\usepackage{algorithm}
\usepackage{algpseudocode}
\usepackage{multirow} 
\usepackage{rotating}
\usepackage{cite}
\usepackage{color}
\usepackage{tabulary}
\usepackage{ctable}
\usepackage{comment}
\usepackage{silence}

\def\colorchanges{black}

\def\mkcolorsup#1#2{\relax}

\def\bfn#1{\bm{\mathbf{#1}}}    
\DeclareMathOperator*{\argmax}{arg\,max}
\DeclareMathOperator*{\argmin}{arg\,min}
\def\req#1{(\ref{#1})}
\DeclareGraphicsExtensions{.eps}

\def\mathcal#1{\mbox{$\cal{#1}$}}

\def\req#1{(\ref{#1})}

\begin{document}

\def\titulo{Contextual Mixture of Experts: Integrating Knowledge into Predictive Modeling}
\title{\titulo}
%

\author{       
Francisco~Souza,
Tim~Offermans,
Ruud~Barendse,
Geert Postma,
and Jeroen Jansen%
 \thanks{© 2022 IEEE.  Personal use of this material is permitted.  Permission from IEEE must be obtained for all other uses, in any current or future media, including reprinting/republishing this material for advertising or promotional purposes, creating new collective works, for resale or redistribution to servers or lists, or reuse of any copyrighted component of this work in other works}
 \thanks{F.\ Souza, T.\ Offermans, G.\ Postma and J.\ Jansen are with the Radboud University, Institute for Molecules and Materials, Heyendaalseweg 135, 6525AJ Nijmegen, The Netherland. Emails: \footnotesize\{f.souza,~t.offermans,~g.j.postma,~jj.jansen\}@science.ru.nl.}
}

\markboth{IEEE Transactions on Industrial Informatics, ~Vol.~X, No.~Y, Month~Year}%
{Souza et. al: 
 \MakeLowercase{\textit{et al.}}: 
\titulo}



\maketitle

\begin{abstract}
This work proposes a new data-driven model devised to integrate process knowledge into its structure to increase the human-machine synergy in the process industry. The proposed Contextual Mixture of Experts (cMoE) explicitly uses process knowledge along the model learning stage to mold the historical data to represent operators’ context related to the process {\color{\colorchanges} through possibility distributions}. This model was evaluated in two real case studies for quality prediction, including a sulfur recovery unit and a polymerization process. The contextual mixture of experts was employed to represent different contexts in both experiments. The results indicate that integrating process knowledge has increased predictive performance while improving interpretability by providing insights into the variables affecting the process’s different regimes.
\end{abstract}



%
\IEEEpeerreviewmaketitle

\section{Introduction}
\label{s.intro}
There is an increasing demand for industrial digitization toward a more sustainable and greener industrial future. Artificial intelligence (AI) is at the front of the 4th industrial revolution by redefining decision-making at the operational and technical levels, allowing faster, data-driven, and automatic decision-making along the value chain. Also, with further growth in industrial data infrastructure, many companies are implementing data-driven predictive models to improve energy efficiencies and industrial sustainability. This can reduce production costs and environmental impact while increasing process efficiency. In parallel, the new upcoming industrial revolution (Industry 5.0) aims to leverage human knowledge and decision-making abilities by strengthening the cooperation between humans and machines \cite{Nahavandi2019}. This new revolution will require the data-driven models to be explainable by providing insights into the process to gain the operator’s trust and increase synergy. The human-machine synergy can be further enhanced by incorporating operator domain knowledge and process information into the data-driven models.

Process information can come from various sources, including first-principle equations and process-specific characteristics \cite{Reis2019}, such as process division structure or multiple operating modes. First-principle models can be combined with data-driven models within the hybrid AI models framework \cite{Sansana2021} or within the informed machine learning framework \cite{Rueden2021}. For the process division structure, the multiblock modeling is a common approach for retaining the explainability of multi-stage processes within a multiblock representation \cite{Reis2019}. Multiple operating modes can be caused by a change in feedstock, operation, seasonality (aka. multimode processes), or the sequence of phases comprising a batch cycle production (aka. multiphase processes) \cite{Kai2020}. These modes can be represented in a multi-model (ensemble) structure \cite{Souza2021}. {\color{\colorchanges} This work focuses on modeling processes with multiple operating modes, and the proposed method was created with this in mind. However, the proposed method is flexible enough to be applied to other types of processes where process expert knowledge is available.}

The modeling of multiple operating modes processes follows from rule-based expert systems \cite{Facco2010,Zhao2014,He2018,Shi2020}, clustering \cite{Luo2016}, Mixture models (MM) \cite{Yu2009,Ge2020,Bei2020}, Gaussian mixture regression (GMR) \cite{Shao2018,Wang2018}, or mixture of experts (MoE) \cite{Souza2014a,Souza2021}
strategies to identify the groups that represent each operational regime, then combine them according to the process's regime. Apart from rule-based expert models, none of the above works discuss using domain knowledge from operators.
In practice, such methods do not attempt to represent process characteristics; rather, the goal is to minimize prediction error, and in some cases, domain knowledge is used only to initialize the model structure. As a result, despite being accurate, the model becomes unrepresentative of the process, making it impossible to understand the effects of the variables in the various regimes of the process. In fact, the rich process context available from operators can be valuable to the model. If a model can reflect the operator's knowledge, it can play an important role in model acceptance in the industry.

This paper proposes the contextual mixture of experts (cMoE), a new data-driven model that connects the process' expert domain knowledge (process context) to the predictive model. The cMoE gives a holistic perspective to the data-driven model while still adhering to process correlation and representing the operators' process context. The cMoE is composed of a set of expert and gate models, where each expert/gate is designed to represent a specific context of the process {\color{\colorchanges} employing possibility distributions}. The gates represent operators’ process context in the model by defining the boundaries of each contextual region. These three components, experts and gates, and the operator’s contextual information form the basis of cMoE. The training procedure in cMoE uses a learning approach devised to assure that each expert represents the defined context and, at the same time, generalizes well for unseen data. To allow model interpretability, the gates and expert models are linear models. Also, a $\ell_1$ regularization penalty is applied to the gates and experts learning for a parsimonious representation of each context model.

{\color{\colorchanges}
The application of regularization in MoE, with linear base models is not new, and it has been used for dealing with high-dimensional settings and for feature selection. For example, \cite{Khalili2010} investigates the $\ell_1$ penalty and smooth clipped absolute deviation (Scad) \cite{Fan2001} penalties for feature selection in MoE. In addition,  \cite{Peralta2014} proposed using the $\ell_1$ penalty for MoE in classification applications. The authors in \cite{Chamroukhi2018,Nguyen2020} investigate the theoretical aspect of MoE with $\ell_1$ penalties. In order to avoid instability in the learning of gating coefficients (typically followed by a softmax function), the authors in \cite{Huynh2019} proposed the use of a proximal-Newton expectation-maximization (EM). The elastic-net penalty was used by the authors of \cite{Souza2021}, who employed a regularization penalty for the inverse of Hessian matrix along the gates learning. 
In the proposed cMoE, the $\ell_1$ penalty is employed to promote sparsity in the gates and experts. The solution for the gates and experts follows from the coordinate gradient descent together with the EM framework, as used in \cite{Souza2021}. However, there are two significant differences between the method presented here and the work of \cite{Souza2021}. The gates and experts models are chosen based on an estimator of the leave-one-out cross-validation error (LOOCV).
The cMoE's performance is accessed based on an estimated LOOCV, and this is used in the EM algorithm as a stop condition and for model selection. Unlike \cite{Souza2021}, where regularized Hessian can lead to unstable results, the learning of the gates in the cMoE is based on the Newton update, with a step-size parameter added to control the learning rate and increase model stability. 
}

The proposed method is evaluated in two experiments. The first one is the sulfur recovery unit (SRU) described in \cite{Fortuna2007}, where the goal is to predict the H\(_2\)S at the SRU's output stream. The operators in that study are more interested in the H\(_2\)S peaks as they are related to the undesirable behavior of SRU unit. The cMoE is then applied for predicting H\(_2\)S with separate representations for peaks and non-peaks components, allowing the identification of causes of the peaks, beyond the prediction of H\(_2\)S. The second case study investigates the application of cMoE to estimate the acid number in a multiphase batch polymerization process. The process knowledge is presented in an annotated data source for the phase transition. The cMoE is then utilized to provide a contextual model for each phase. In all the experiments, the results indicate that incorporating the operator’s context into the cMoE gives interpretability and insight to the process and significantly increases the model performance. 

{\color{\colorchanges}
This work's main contributions are as follows: i) the use of possibility distributions to represent the operator's expert knowledge; ii) the development of a new mixture model called contextual mixture of experts to incorporate the operator's expert knowledge from the possibility distributions into the model structure; iii) the application of a $\ell_1$ penalty to the gates and experts coefficients to promote sparse solutions. iv) the development of a leave-one-out error (LOOCV) estimator for experts and gates and the cMoE model;
}

The paper is divided as follows. Section \ref{s.pre} gives the background for the paper. Section \ref{s.cmoe} presents the proposed model contextual mixture of experts. Section \ref{s.experimental} presents experimental results. Finally, Section \ref{s.conclusions} gives concluding remarks.

\section{\color{\colorchanges} Preliminaries}
\label{s.pre}
\subsection{ Notation}
In this paper, finite random variables are represented by capital letters and their values by the corresponding lowercase letters, e.g.\ random variable $A$, and corresponding value $a$. Matrices and vectors are represented by boldface capital letters, e.g.\ $\bfn{A}=[a_{ij}]_{N\times d}$ and boldface lowercase letters, e.g.\ $\bfn{a}=[a_1,\ldots,a_d]^T$, respectively. The input and output/target variables are defined as $X=\{X_1, \ldots, X_d\}$ and $Y$, respectively. The variables $X_1, \ldots, X_d$ can take $N$ different values as $\{x_{ij} \in X_j \: : \: j=1,\ldots,d \:\:\text{and}\:\: i=1,\ldots,N \}$, and similarly for $Y$ as $\{y_{i} \in Y \: : \: i=1,\ldots,N \}$.

\subsection{\color{\colorchanges} Mixture of Experts}
The MoE is a modeling framework based on the divide and conquer principle. It consists of a set of experts and gates, with the gates defining the boundaries (soft boundaries) and the experts making predictions within the region assigned by the gates. The prediction output of a MoE with $C$ experts is
\begin{align}
    \hat{y}(\bfn{x}_i)= \sum_{c=1}^{C} g_{c}(\bfn{x}_i)\hat{y}_{c}(\bfn{x}_i),
    \label{equ.cmoepred}
\end{align}
where $\hat{y}_{c}(\bfn{x}_i)$ is the  expert's predicted output at region $c$, and $g_{c}(\bfn{x}_i)$ is the gate function that represents the expert's boundaries at region $c$.

The probability distribution function (PDF) of the MoE is defined as
\begin{equation}
p(y_i|\bfn{x}_i;\Omega) = \sum_{c=1}^{C} g_c(\bfn{x}_i;\CMcal{V}) p(y_i|\bfn{x}_i;\Theta_c),
\end{equation}
The PDF is the expert $c$ conditional distribution with mean $\hat{y}_{ci}$, and  $\sigma^2_c$ is the noise variance. 
The set of expert parameters is defined as $\bm{\Theta}_c=\{\bm{\theta}_c,\sigma^2_c\}$. The gates $g_c(\bfn{x}_i;\CMcal{V})$ assigns mixture proportions to the experts, with constraints $\sum_{c=1}^Cg_{c}(\bfn{x}_i)=1$ and $0\leq g_{c}(\bfn{x}_i)\leq 1$, and for simplicity $g_{ci}=g_{c}(\bfn{x}_i)$. The gates typically follows from the softmax function:
\begin{align}
   g_{ci}={\exp\left(\bfn{x}_i^T\bfn{v}_{c}\right)}\Big/{\sum_{k=1}^{C}\exp\left(\bfn{x}_i^T\bfn{v}_{k}\right)}
  \label{equ.gateDef}
\end{align}
where $\bfn{v}_c$ is the parameter that governs the gate $c$, and $\CMcal{V}=\{\bfn{v}_1,\ldots,\bfn{v}_C\}$ is the set of all gates parameters.
The collection of all parameters is defined as ${\Omega}=\{{\Theta}_1,\ldots,{\Theta}_C,\bfn{v}_1,\ldots,\bfn{v}_C\}$. 

From the MoE framework, the parameters in $\Omega$ are found trough the maximization of log-likelihood
\begin{align}
\Omega = \argmax_{\Omega^*} \CMcal{L}({\Omega}^*),
\label{equ.OmegaMaxLike}
\end{align}
where the log-likelihood for $N$ iid samples is defined as
$\CMcal{L}({\Omega}) =  \log p({Y}|{X};\Omega) = \sum_{i=1}^{N} \log p(y_i|\bfn{x}_i;\Omega) $. 
The solution of Eq. \req{equ.OmegaMaxLike} follows from the expectation-maximization (EM) algorithm, an iterative procedure that maximizes the log-likelihood from successive steps.
In EM, the $p({Y}|{X};\Omega)$ is treated as the incomplete data distribution. The missing part, the hidden variables ${Z}$ are introduced to indicate which expert $c$ was responsible to generate the sample $i$. The complete distribution is given by 
\begin{equation}
p(y_i,\bfn{z}_{i}|\bfn{x}_i;\Omega) = \prod_{c=1}^{C}  g_c(\bfn{x}_i;\CMcal{V})^{z_{ci}} p(y_i|\bfn{x}_i;\Theta_c)^{z_{ci}} 
\end{equation}
where $z_{ci} \in \{0,1\}$ and for each sample $i$, all variables $z_{ci}$ are zero, except for a single one. The hidden variable $z_{ci}$ indicates which expert $c$ was responsible of generating data point $i$. 
Let $\hat{\Omega}^t$ denote the estimated parameters at iteration $t$, the EM algorithm increases the log-likelihood at each iteration so that ${\CMcal{L}}_C(\hat{\Omega}^{t+1}) > {\CMcal{L}}_C(\hat{\Omega}^{t})$. It is composed by two main steps, the expectation (E-step) and maximization step (M-step). 

From an initial guess $\hat{\Omega}^{0}$, the expectation of the complete log-likelihood  (aka $Q$-function) is computed with respect to the current estimate $\hat{\Omega}^{t}$:
\paragraph{E-step} $Q^{t}(\Omega) = \text{E}_{}\left[\tilde{\CMcal{L}}^{}_C({\Omega}) \, | \, {Z}, \hat{\Omega}^t\right]$
\begin{align}
Q^{t}(\Omega) &= \sum_{i=1}^{N}\sum_{c=1}^{C} \gamma^{t}_{ci}\log \Big[ g_c(\bfn{x}_i;\CMcal{V}) p(y_i|\bfn{x}_i;\Theta_c)  \Big]^{}.
\end{align}
where $\gamma^t_{ci}\equiv \text{E}\left[z_{ci}\, |\, {Z}, \hat{\Omega}^t\right]$ is the posteriori distribution of $z_{ci}$ after observing the data ${X}$, ${Y}$ (called responsibilities). The responsibilities accounts for the probability of expert $c$ has generated the sample $i$.

In the M-Step, the new parameters are found by maximizing the $Q$-function, as the following
\paragraph{M-step} 
\begin{align}
\hat{\Omega}^{t+1} = \argmax_{\Omega} Q^{t}(\Omega)
\end{align}
The EM runs until convergence, typically measured by the $Q$-function. 
{\color{\colorchanges}

\subsection{Possibility distribution}
Possibility theory is a framework for representing uncertainty, and ambiguous knowledge, from possibility distributions \cite{Zadeh1978,Dubois1998}.
Let $\Psi$ represent a finite set of mutually exclusive events, where the true alternative is unknown. This lack of information on the true event resumes the uncertainty in representing the true alternative. A possibility distribution, defined in the set $\Psi$, maps the set of possible events to the unit interval $[0,1]$, $ \pi\,:\,\Psi  \rightarrow [0,1]$, with at least $\pi(s)=1$ for some ${s} \in \Psi$, where the function $\pi(s)$ represents the state of knowledge by the expert about the state of the data,
and $\pi(s)$ stands as the belief of event $s$ be the true alternative. The larger $\pi(s)$, the more plausible, i.e. plausible the event $s$ is. It can also be interpreted as the ``degree of belief''; for example, the ``degree of belief'' of event $s$ be true is $0.8$. It is assumed to be possibilistic rather than probabilistic, so distinct events may simultaneously have a degree of possibility equal to $1$. 

Given two possibilities distributions $\pi_{1}(s)$ and $\pi_{2}(s)$, the possibility distribution $\pi_{1}(s)$ is said to be more specific than  $\pi_{2}(s)$, iff $\pi_{1}(s)<\pi_{2}(s)$  $\forall s \in A$. Then, $\pi_1$ is at least as restrictive and informative as $\pi_{2}$. In the possibilistic framework, extreme forms of partial knowledge can be captured, namely:
\begin{itemize}
    \item {\em Complete knowledge}: for some $s_0$, $\pi(s_0)=1$, and {\color{\colorchanges}$\pi(s)=0, \, \forall s\neq s_0$} (only $s_0$ is possible);
    \item {\em Complete ignorance}: $\pi(s)=1\,, \forall s\in  \Psi$ (all events are possible).
\end{itemize}
The more specific $\pi$ is, the more informative it is. The minimal specificity principle drives possibility theory \cite{Yager1982}. It states that any now-known impossible hypothesis cannot be ruled out. It is a principle of minimal commitment, caution, and information. Essentially, the aim is to maximize possibility degrees while keeping constraints in mind.

}

\section{Contextual Mixture of Experts}
\label{s.cmoe}
In this section the contextual mixture of experts is presented. The first subsection Sec. \ref{ss.model} describes the model structure and its goals. {\color{\colorchanges} Sec. \ref{ss.expertknowledge} introduces the possibilities distributions used in the expert knowledge representation}. The learning of the contextual mixture of experts is given in Sec. \ref{s.learning}. \vspace{-1em}

\subsection{The Model}
\label{ss.model}
\begin{figure}[!t]
\begin{center}
\includegraphics[width=.8\columnwidth]{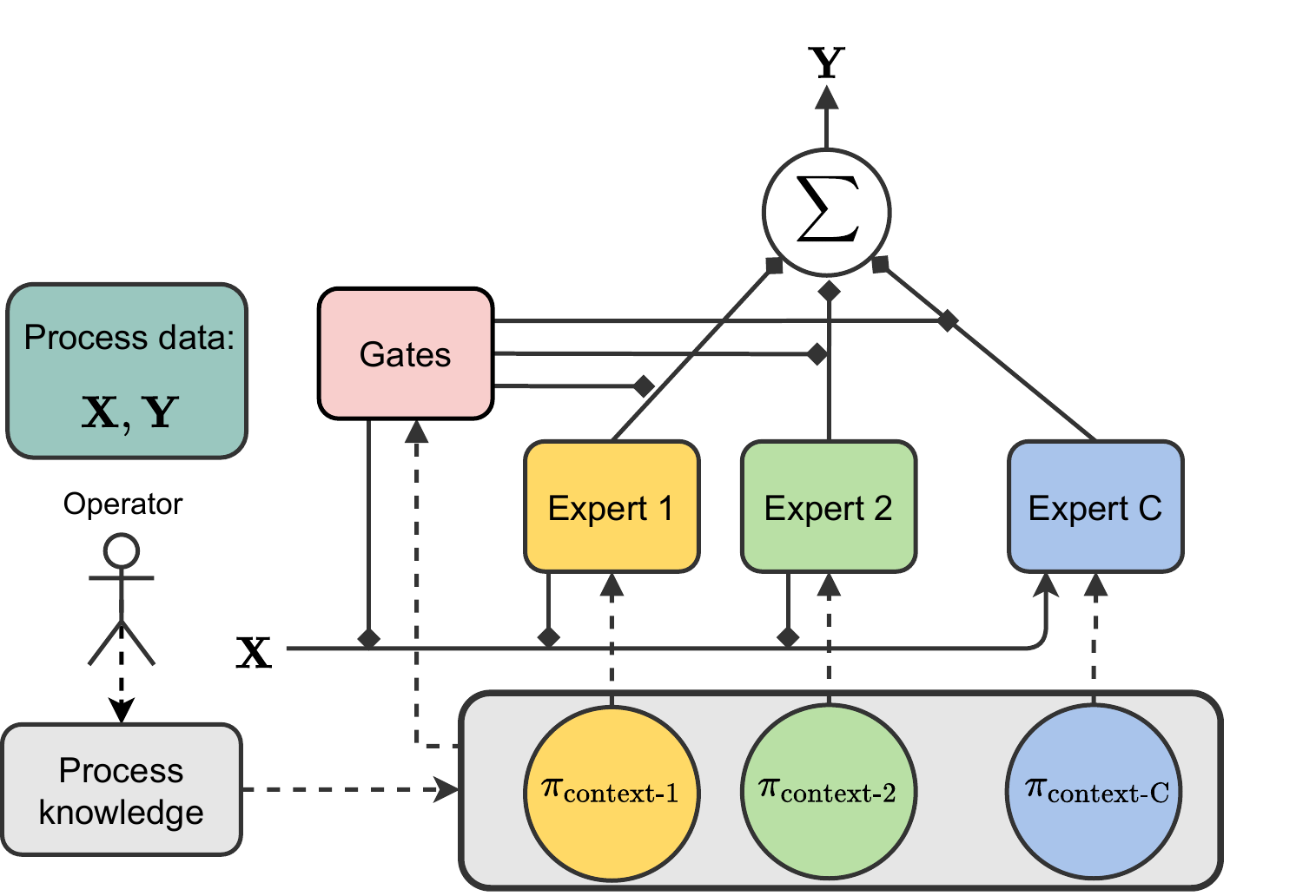}
\caption{{\color{\colorchanges}MoE representation with $C$ experts. Solid lines indicates direct data flow, while dashed lines indicates flow of expert knowledge. The process knowledge is encoded via the possibility distributions 
$\{ \pi_{\text{context-1}},\ldots,\pi_{\text{context-C}} \}$ }. }
\label{fig.cMoE} 
\vspace{-1.7em}
\end{center}
\end{figure}

The structure of the contextual mixture of experts is composed by $C$ experts, and gates models; this is represented in Fig. \ref{fig.cMoE}. 
\relax
The context here refers to any meaningful process data characteristic defined by the analyst or derived from any process information/knowledge. {\color{\colorchanges} Each context $c$ is encoded by a possibility distribution $\pi_c$, which is used to represent the expert/analyst uncertainty knowledge about the respective context.}
The analyst inputs each context into the contextual mixture of experts by incorporating the context into the model structure and defining each expert model's expected operating region.
Then, each contextual expert model $\hat{y}_{c}(\bfn{x}_i)$ is trained on the region of context $c$ and makes predictions based on its domain representation. This contextual approach enables the prediction to be divided into different components representing meaningful context specified by the analyst.

The output prediction of cMoE is given by a weighted sum of the experts output, {\color{\colorchanges} given by Eq. \req{equ.cmoepred}}. 
In cMoE the the gates $g_{c}(\bfn{x}_i)$ is the probability of sample $\bfn{x}_i$ belonging to the region of context $c$. The expert model $\hat{y}_{c}(\bfn{x}_i)$ is trained on the region defined by the context $c$, and gives the prediction according with its domain representation. For simplified notation, the contribution of each expert model is defined as
$\hat{y}_i = \sum_{c=1}^{C} h_{ci}$
where the input $\bfn{x}_i$ is omitted for clarity, and $h_{ci} \triangleq  g_{c}(\bfn{x}_i)\hat{y}_{c}(\bfn{x}_i)$, {\color{\colorchanges} {\color{\colorchanges} where $\triangleq $ stands for defined to be}}. For example, in a 3-phase batch process, the cMoE is set to have three contexts, each one representing a phase. In such case, the simplified representation is given by
$$\hat{y} = h_{\text{phase}_1} + h_{\text{phase}_2} + h_{\text{phase}_3}$$
From that, each contextual model can be interpreted separately, or jointly according to the analyst's needs.

\subsection{\color{\colorchanges} Expert Process Knowledge Representation}
\label{ss.expertknowledge}
{ \color{\colorchanges}
Here, the expert's knowledge for each context is represented by a possibility distribution. For each context $c$, there is an associated possibility distribution $\pi_c(\bfn{x}_i)$ (in short $\pi_{ci}$), where $i \in \Psi$ (where $\Psi$ represents the set of all available samples). The value of $\pi_{ci}$ indicates the degree of belief of the sample $i$ pertains to the region of context $c$. In the case of $\pi_{ci}=1$, sample $i$ is considered to be fully possible; if $\pi_{ci}=0$, sample $i$ is considered to be completely impossible to be part of context $c$; any value between these two extremes $\pi_{ci}=p, \,\, p\in]0,1[$, can be accredited as partial possibility, to a degree certainty $p$, of sample $i$ belonging to the context $c$. Therefore, if $\pi_{ci}=1$ for all $c=1,\ldots,C$, the sample $i$ will be defined as being believed to belong to all contextual regions ({\em complete ignorance} about sample $i$), whereas if $\pi_{ci}=1$ for a single $c \in \{1,\ldots,C\}$, the sample $i$ will be defined as be being accredited to to be fully certain to belong to context $c$ ({\em complete knowledge} about sample $i$), while being impossible to belong the other contexts.

Because expert knowledge's reliability cannot be fully guaranteed for each context, and because reliability is commonly described with some degree of certainty, for example (80\% sure or 70\% certain) \cite[Chapter~2]{Solaiman2019}, the possibility distributions used here are intended to account for this uncertainty in representing expert knowledge of each context.
To do so, two possibility distributions are employed in the experiments, the $\alpha$-Certain distribution and the $\beta$-Trapezoidal (fuzzy) distribution, where $\alpha$ and $\beta$ are the degrees of certainty.

\paragraph{$\alpha$-Certain possibility distribution} is an imprecise knowledge distribution with a certainty factor $\alpha$. The available knowledge about the true alternative is expressed as a subset $A\subseteq \Psi$ associated with a certain level of trust $\alpha \in [0,1]$, concerning the occurrence of $A$. This can be expressed declaratively as ``$A$ is certain to degree $\alpha$''.
    This type of distribution has been suggested as \cite{DUBOIS2014}:
    \begin{equation}
    \pi_{ci} =
    \begin{cases}
      1 & \text{if $i \in  A$}\\
      1-\alpha & \text{otherwise}\\
    \end{cases} 
\end{equation}
If $\alpha=1$, $\pi_{ci}$ represents the characteristic function of $A$, on the other extreme if $\alpha=0$, $\pi_{ci}$ represents the total ignorance about $A$. The $\alpha$-Certain possibility distribution is shown in Fig. \ref{fig.possibility_alpha_certain}a.
    \begin{figure}[!t]
   \begin{center}
   \includegraphics[width=1\columnwidth]{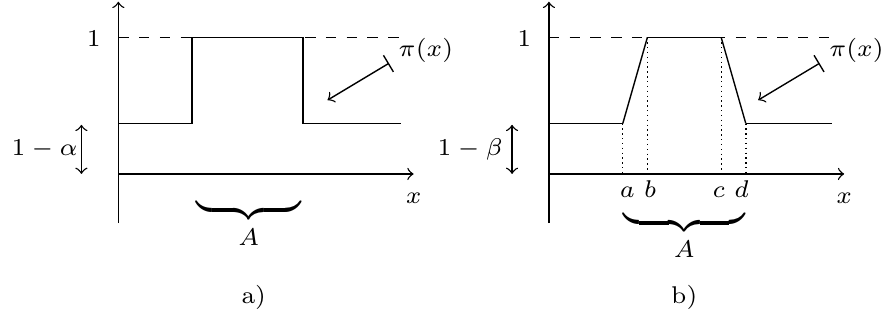}
   \caption{Possibilistic distributions, a) $\alpha$-Certain possibility distribution, b) $\beta$-Trapezoidal epistemic possibility distribution.}
 \label{fig.possibility_alpha_certain}
   \vspace{-1.8em}
   \end{center}
 \end{figure}\relax

\paragraph{$\beta$-Trapezoidal epistemic possibility distribution} In the epistemic (fuzzy like) possibility distribution, each event has a degree of belief associated with the possibility of an event. In the epistemic distribution, the available knowledge about the true alternative is given as a constrain defined in terms of ``a fuzzy concept'' defined in $\Psi$. All standard types defining membership function and representing fuzzy constraints (i.e., triangular, trapezoidal, Gaussian, etc.) can be applied to define epistemic type possibility distributions. In here, the trapezoidal function is defined by a lower limit $a$, an upper limit $d$, a lower support limit $b$, and an upper support limit $c$, where $a < b < c < d$:
        \begin{equation}
    \pi_{ci} =\max\left(\min\left(\frac{i-a}{b-a},1,\frac{d-i}{d-c}\right),\beta\right)
\end{equation}
To account for the unreliable, a certainty factor $\beta \in [0,1]$ is added, where $\beta=0$ means a fully unreliable source, and $\beta=1$ means a fully reliable source. the $\beta$-Trapezoidal distribution is shown in Fig.\ \ref{fig.possibility_alpha_certain}b.
More possibility distributions are also fited in this framework; for more possibilities distributions, see \cite[Chapter~2]{Solaiman2019}.
\vspace{-.3em}
}

\subsection{The Learning}
\label{s.learning}

{\color{\colorchanges} The goal of cMoE is to integrate the expert knowledge representation (via the possibility distribution) into the model structure. To constrain the contextual information, the} parameters in $\Omega$ are found trough the maximization of the weighted log-likelihood (WLL) of $\Omega$, defined as %
\begin{align}
\CMcal{L}_C({\Omega}) = \sum_{i=1}^{N} 
\log p(y_i|\bfn{x}_i;\Omega)^{\bfn{\pi}_i}
\end{align}
where $\bfn{\pi}_i=[\pi_{1i},\ldots,\pi_{ci}]^T$ is the contextual weight vector {\color{\colorchanges}from the expert knowledge} and must be specified a {\em priori}. 
The r.h.s power is defined as 
$p(y_i|\bfn{x}_i;\Omega)^{\bfn{\pi}_i} \triangleq \sum_{c=1}^{C} \Big[ g_c(\bfn{x}_i;\CMcal{V}) p(y_i|\bfn{x}_i;\Theta_c) \Big]^{\pi_{ci}}.$
%
The weighted ML estimation of $\Omega$ constrains the contextual information into the model structure, laying down the basis of cMoE framework.
{\color{\colorchanges}The idea is to down-weight samples that have a low degree of belief to belong to regions of expert $c$}.

The maximization of the WLL also $\CMcal{L}_C({\Omega})$ follows from the EM algorithm. 
By inserting the contextual weights, the WLL of the complete data distribution becomes 
\begin{equation}
\tilde{\CMcal{L}}^{}_C({\Omega}) = \sum_{i=1}^{N} \log p(y_i,\bfn{z}_{i}|\bfn{x}_i;\Omega)^{\bfn{\pi}_i},
\end{equation}
where the power at the r.h.s is defined as 
\begin{equation}
p(y_i,\bfn{z}_{i}|\bfn{x}_i;\Omega)^{\bfn{\pi}_i} \triangleq   \prod_{c=1}^{C} \Big[ g_c(\bfn{x}_i;\CMcal{V})^{z_{ci}} p(y_i|\bfn{x}_i;\Theta_c)^{z_{ci}}  \Big]^{\pi_{ci}}. \nonumber
\end{equation}
Let $\hat{\Omega}^t$ denote the estimated parameters at iteration $t$ of the EM algorithm. {\color{\colorchanges} The expectation (E-step) and maximization steps (M-step) for WLL are:} 
\paragraph{E-step} From an initial guess $\hat{\Omega}^{0}$, the expectation of the complete WLL (aka $Q$-function) is computed with respect to the current estimate $\hat{\Omega}^{t}$:
\begin{align}
Q^{t}(\Omega) 
    &= \sum_{i=1}^{N}\sum_{c=1}^{C} \pi_{ci}\gamma^{t}_{ci}\log \Big[ g_c(\bfn{x}_i;\CMcal{V}) p(y_i|\bfn{x}_i;\Theta_c)  \Big]^{}.
\end{align}
where $$\gamma^t_{ci} =  \frac{\pi_{ci}g_c(\bfn{x}_i;\CMcal{V}^t) p(y_i|\bfn{x}_i;\Theta^t_c)}{\sum_{k=1}^{C}  \pi_{ki}g_k(\bfn{x}_i;\CMcal{V}^t) p(y_i|\bfn{x}_i;\Theta^t_k) } $$
$\gamma^t_{ci}$ are the responsibilities of expert $c$ generated sample $i$. {\color{\colorchanges}It should be noted that the responsibilities are also a result of the contextual weights, in the case of $\pi_{ci}=0$, the responsibility of expert $c$ is $\gamma^t_{ci}=0$, indicating that context $c$ has no role in generating the sample $i$.}

\paragraph{M-step} In the M-Step, the new parameters are found by maximizing the $Q$-function, as the following
\begin{align}
\hat{\Omega}^{t+1} = \argmax_{\Omega} Q^{t}(\Omega)
\end{align}
The $Q$-function is further decomposed to account for the gates and experts contributions separately, as
$Q^{t}(\Omega) =  Q^{t}_g(\CMcal{V}) + Q^{t}_e(\Theta) $,
where
\begin{align}
Q^{t}_g(\CMcal{V}) &= \sum_{i=1}^{N}\sum_{c=1}^{C}  \pi_{ci}\gamma^{t}_{ci}\log g_p(\bfn{x}_i;\CMcal{V})\nonumber \\
Q^{t}_e(\Theta) &=\sum_{i=1}^{N}\sum_{c=1}^{C} \pi_{ci}\gamma^{t}_{ci}\, \log p(y_i|\bfn{x}_i;\Theta_c)  \nonumber 
\end{align}
The maximization is performed separately for the experts and gates in the updating phase. Here, the experts and gates models are linear, despite more complex models being allowed and easily integrated into this framework.
\vspace{-1em}
\subsection{Experts Learning}
In the maximization step, the updated experts parameters are found from the maximization of $Q^t_e({{\Theta}}^{})$. The contribution of the individual experts can be accounted separately:
\begin{align}
 Q^t_e({{\Theta}}^{}) = \sum_{c=1}^C  Q^t_{ec}({\bfn{\theta}_c}^{}) =  \sum_{c=1}^C\sum_{i=1}^N \pi_{ci} \gamma^t_{ci} \log \CMcal{N} \left(y_i| \bfn{x}^T_i\bfn{\theta}_c,\sigma_c^2\right)
\end{align}
where $Q^t_{ec}(\cdot)$ accounts for the contribution of expert $c$. Hence, the parameters of expert $c$ can be updated apart from the other experts. Equivalently, instead of maximizing $Q^t_{ec}(\bfn{\theta}_c^{})$, the updated coefficient is found by minimizing the negative of $Q^t_{ec}(\bfn{\theta}_c^{})$, as 
\begin{align}
\hat{\bfn{\theta}}_c^{t+1} 
&=\argmin_{\bfn{\theta}_c} \left(\frac{1}{2} \sum_{i=1}^N \pi_{ci} \gamma^t_{ci} \left(y_i - \bfn{x}^T_i\bfn{\theta}_c\right)^2 \right). \label{equ.max_theta_p_en}
\end{align}
To promote the sparsity of expert $c$ coefficients, a $\ell_1$ penalty is added to Eq. \req{equ.max_theta_p_en}, leading to 
\begin{align}
\small
\hat{\bfn{\theta}}_c^{t+1} \! =\argmin_{\bfn{\theta}_c} \left(\frac{1}{2} \sum_{i=1}^N \pi_{ci} \gamma^t_{ci} \left(y_i - \bfn{x}^T_i\bfn{\theta}_c\right)^2 + \lambda_c^e \sum_{j=1}^d|\theta_{cj}| \right), \label{equ.min_theta_p_en}
\end{align}
where $\lambda_c^e$ controls the importance of the regularization penalty. This penalty, also known as least absolute shrinkage and selection operator (LASSO), drives irrelevant features coefficients towards zero. This characteristic is suitable for industrial applications where not all variables are relevant to the prediction, providing compact models. Under the cMoE, this penalty will allow the selection of parsimonious models for each expert, reducing the complexity of the overall model structure.  Together with the contextual information, the LASSO penalty will provide a relevant set of features for each expert domain, thus allowing a better interpretation of the model representation, as well allowing the learning in scenarios with small number of samples and many features.

The minimization of Eq. \req{equ.min_theta_p_en} will follow from the coordinate gradient descent (CGD). In CGD, each coefficient is minimized individually at a time. 
The updated regression coefficient of variable $j$ and expert $c$ is given as 
\begin{equation}
\hat{\theta}^{t+1}_{cj}  = {S\left(\sum_{i=1}^N  \pi_{ci}\gamma^t_{ci} x_{ij} (y_{i} - \tilde{y}_{ci}^j ),\lambda^e_c \right)}\Big/{\sum_{i=1}^N   \pi_{ci}\gamma^t_{ci} x_{ij}^2 },
\label{equ.cgd_expert_update}
\end{equation}
where $\tilde{y}_{ci}^j = \sum_{l\neq j}^d x_{il}\theta^t_{cl}$ is the fitted value of local expert $c$, without the contribution of variable $j$ and the $S(z,\eta)$ is the soft threshold operator, given by $S(z,\eta)=\text{sign}(z)(|z|-\eta)_+$.
From Eq. \req{equ.cgd_expert_update} the contextual weight adds a weighting factor over the responsibilities. In the case where $\pi_{ci}=1$ for all experts, the responsibility will be the primary driving force in determining the contribution for that specific sample. 

\subsubsection{Experts Model Selection}
\label{sss.experts_selection}
The selection of LASSO regularization can follow from the $k$-fold cross-validation ($k$-CV) error. However, the $k$-fold cross-validation for the LASSO may present potential bias issues in high-dimensional settings. The reason for bias is that the data change caused by the splitting into folds affects the results. Allowing $k$ to be large enough to reduce the bias is one possible solution. Choosing $k=N$, for example, results in the leave-one-out cross-validation (LOOCV) error, an unbiased estimator for the LASSO error. However, the computation of LOOCV is heavy as it requires training the model $N$ times, and one solution is to approximate the LOOCV from the data. Under mild conditions assumptions \cite{Stephenson2020}, the prediction LOOCV of linear models with LASSO penalty can be approximated from its active set; the active set is the index of those variables with nonzero coefficients.

Let the active set of the expert $c$ be $\CMcal{E}^{}_{c}=\{j \in \{1,\ldots,d\} \:| \: \theta_{cj} \neq  0\}$. Also, let $\bfn{X}_{\CMcal{E}^{}_{c}}$ for the columns of matrix $\bfn{X}$ in the set $\CMcal{E}^{}_{c}$. 

The LOOCV of expert $c$ can be approximated by
\begin{equation}
\label{e.loocv_expert}
 {\text{CV}}^{e}_{c}(\lambda^e_p)=\sum_{i=1}^{N} \pi_{ci}\gamma^t_{ci}\left(y_i - \hat{y}^{(-i)}_{ci}\right)^2
\end{equation}
where $\hat{y}^{(-i)}_{ci}$ is the estimated output of expert $c$ without sample $i$. The estimation of $\hat{y}^{-i}_{ci}$ from the active set is given by
\begin{equation}
 \hat{y}^{(-i)}_{ci} = \frac{(\hat{y}_{ci} - [\bfn{H}_{c}]_{ii}y_i)}{1-[\bfn{H}_{c}]_{ii}}.
 \label{e.y_minus_i}
\end{equation}
where $[\bfn{H}_{c}]_{ii}$ are the $i$th diagonal element of hat-matrix of expert $c$, which is defined as 
\begin{equation}
\bfn{H}_c = \bfn{X}_{\CMcal{E}_{c}}(\bfn{X}^T_{\CMcal{E}_{c}}\bfn{\Gamma}_c\bfn{X}_{\CMcal{E}_{c}})^{-1}\bfn{X}^T_{\CMcal{E}_{c}}\bfn{\Gamma}_c\nonumber
\end{equation}
where $\bfn{\Gamma}_c=\text{diag}( \pi_{c1}\gamma^t_{c1},\ldots, \pi_{cn}\gamma^t_{cn})$ is the diagonal matrix of contextual weights and responsibilities, and $\bfn{X}_{\CMcal{E}_{c}}$ is the active set subset of matrix $\bfn{X}$. 
{\color{\colorchanges} In here, the inverse $(\bfn{X}^T_{\CMcal{E}_{c}}\bfn{\Gamma}_c\bfn{X}_{\CMcal{E}_{c}})^{-1}$ is computed via the \ LU decomposition. If this is not possible, e.g.\ due the matrix becoming too large, one could estimate the validation error from an independent validation set.}
The value of $\lambda^e_p$ selected is the one that minimizes the value of ${\text{CV}}^{e}_{M}$. 
Note that this estimator resembles the predicted residual sum of squares (PRESS), except that only the active set is used along with the LOOCV estimation for the LASSO.

\subsection{Gates Learning}
For the gates, the new updated parameters results from the maximization of $Q^t_g({\CMcal{V}}^{})$, or equivalently by minimizing $-Q^t_g({\CMcal{V}}^{})$. By expanding the gates contribution, it becomes
\begin{align}
\small
Q^t_g({\CMcal{V}}^{}) & = 
 \sum_{i=1}^N \left[\sum_{c=1}^{C} \pi_{ci} \gamma^t_{ci}\bfn{x}_i^T\bfn{v}_c - \phi_i\log \left(\sum_{k=1}^C\exp\left(\bfn{x}_i^T\bfn{v}_k\right)\right)\right], \label{equ.Qtg}
\end{align} 
where $\phi_i=\sum_{c=1}^{C} \pi_{ci} \gamma^t_{ci}$. The function $Q^t_g({\CMcal{V}}^{})$ is concave in the parameters, and its maximization of $Q^t_g({\CMcal{V}}^{})$ will follow the Newton's method. Let $\{\hat{\bfn{v}}_{c}^{t}\}_{c=1}^{C}$ be the current estimates of gates coefficients, the second-order (quadratic) Taylor approximation of Eq. \req{equ.Qtg} around $\{\hat{\bfn{v}}_{c}^{t}\}_{c=1}^{C}$ is:
\begin{align}
\tilde{Q}^t_g({\CMcal{V}}^{}) &= \sum_{c=1}^{C}  Q^t_{gc}(\bfn{v}_c^{}) + C(\{\hat{\bfn{v}}_{c}^{t}\}_{c=1}^{C})\\
Q^t_{gc}(\bfn{v}_c^{}) &= -\frac{1}{2}\sum_{i=1}^N r_{ci} \left(z_{ci} - \bfn{x}^T_i\bfn{v}_c \right)^2,\label{equ.Qgp}
\end{align}
where $\tilde{Q}^t_g({\CMcal{V}}^{})$ is the second order Taylor approximation of $Q^t_g({\CMcal{V}}^{}) $, $Q^t_{gc}(\bfn{v}_c^{})$ accounts by the individual contribution of gate $c$, and $C(\{\bfn{v}_{c}^{t}\}_{c=1}^{C})$ is a constant term, while $r_{ci}$ and ${z}_{ci}$ are given by
\begin{align}
r_{ci} &= \phi_i g_{ci}(1-g_{ci}) , \label{equ.rci}\\
{z}_{ci} &= \bfn{x}^T_i{\bfn{v}}^t_c + \eta \frac{ \pi_{ci} \gamma^t_{ci}-\phi_ig_{ci}}{\phi_ig_{ci}(1-g_{ci})}, \label{equ.zci}
\end{align}
with the gates $g_{ci}$ computed from Eq. \req{equ.gateDef}, and the parameter $\eta$ is a magic parameter added to control the Newton update on the optimization phase.

By adding the LASSO penalty to the gates contribution Eq. \req{equ.Qgp}, the new gate coefficients are found as the solution of the following minimization problem
\begin{equation}
\hat{\bfn{v}}_{c}^{t+1} = \argmin_{{\bfn{v}_c}^{}} \left[\frac{1}{2}\sum_{i=1}^N r_{ci} \left(z_{ci} - \bfn{x}^T_i\bfn{v}_c \right)^2 + \lambda_c^g \sum_{j=1}^d|v_{cj}|  \right].\nonumber
\label{equ.min_gates}
\end{equation}
The gate coefficients are updated from successive local Newton steps. It cycles trough all $C$ gates sequentially, where the values of $g_{ci}$ are calculated from $\{\hat{\bfn{v}}_{c}^{t}\}_{c=1}^{C}$, and they must be updated as soon a new ${\hat{\bfn{v}}}_{c}^{t}$ is computed. The computation of Eq. \req{equ.min_gates} must be repeated until the coefficients converge; usually, few iterations (less than $10$) are needed to the reach convergence.

The solution of Eq. \req{equ.min_gates} is achieved from the CGD, in which
\begin{equation}
\hat{{v}}^{t+1}_{pj}  = {S\left(\sum_{i=1}^N r_{ci} x_{ij} ({z}^{}_{ci} - \tilde{z}_{ci}^j ),\lambda^g_c \right)}\Big/{\sum_{i=1}^N  r^{}_{ci} x_{ij}^2  }
\label{equ.cgd_gate_update}
\end{equation}
where $\tilde{z}_{ci}^j = \sum_{l\neq j}^d z_{il}v^t_{cl}$ is the fitted value of gate $c$, without considering variable $j$.

\paragraph{Practicalities in gates update} Along the gates coefficients update, some practical issues must be taken
\begin{itemize}
    \item Care should be taken in the update of Eq. \req{equ.zci}, to avoid coefficients diverging in order to achieve fitted $g_{ci}$ of $0$ or $1$. When $g_{ci}$ is within $\xi = 10^{-3}$ of $1$, Eq. \req{equ.zci} is set to ${z}_{ci} = \bfn{x}^T_i{\bfn{v}}^t_c$, and and the weights in \req{equ.rci} $r_{ci}$ are set to $\xi$. $0$ is treated similarly.
    \item The use of full Newton step $\eta=1$ in optimization Eq. \req{equ.min_gates} do not guarantee the converge of coefficients \cite{Lee2006}. To avoid this issue, along the experiments $\eta$ was fixed to $\eta=0.1$.
\end{itemize}

\subsubsection{Gates Model Selection}
Similar to the experts’ procedure, the gates model selection will follow from the estimated LOOCV error. The predicted gate output without the sample $i$ is given by
\begin{equation}
 \hat{z}^{(-i)}_{ci}={(\hat{z}_{ci} + [\bfn{M}_{c}]_{ii}z_i)}\Big/{\left(1-[\bfn{M}_{c}]_{ii}\right)}
\end{equation}
where $\bfn{M}_{c}$ is the gate hat matrix at each step of the Newton update, and is computed as
\begin{equation}
 \bfn{M}_c = \bfn{X}_{\CMcal{G}_{c}}( \bfn{X}^T_{\CMcal{G}_{c}}\bfn{R}_c\bfn{X}_{\CMcal{G}_{c}} )^{-1}\bfn{X}^T_{\CMcal{G}_{c}}\bfn{R}_c
\end{equation}
where $\CMcal{G}_{c}=\{j \in \{1,\ldots,N\} \:| \: v_{cj} \neq  0\}$ is the active set of gate $c$, and $\bfn{R}_c =\text{diag}( r_{p1},\ldots, r_{pn})$. The LOOCV is then estimated as
\begin{equation}
\label{e.loocv_gate}
 {\text{CV}}^{g}_{c}(\lambda^g_c)=\sum_{i=1}^{N} r_{ci}\left(z_i - \hat{z}^{(-i)}_{ci}\right)^2
\end{equation}
The gate regularization parameter is selected to minimize the estimated LOOCV error.

\subsection{EM Stop Condition}
\label{sss.modelselection_experts}
In cMoE, the information on the number of experts must be known a $priori$, or can be defined by the analyst, so this is not an issue for the design. The EM algorithm's condition stops must be properly defined to avoid overfitting or a poorly chosen model. Because the implementation is based on a set of linear models, an approximation of LOOCV error is used to assess the model's quality and set the EM algorithm's stop criteria. The estimated LOOCV for cMoE is given by
\begin{align}
\small
 \text{CV}({\Omega}) =\frac{1}{N}\sum_{i=1}^N \left(y_i - \sum_{c=1}^{C} g_{ci}^{(-i)} \hat{y}^{(-i)}_{ci}\right)^{2}
 \label{e.loocverror}
\end{align}
where $\hat{y}^{(-i)}_{ci}$ is given by Eq. \req{e.y_minus_i} and $g_{ci}^{(-i)}$ is given by
\begin{align}
\small
   g^{(-i)}_{ci}={\exp\left({z}^{(-i)}_{ci}\right)}\Big /{\sum_{k=1}^{C}\exp\left(\hat{z}^{(-i)}_{ki}\right)}.
\end{align}
%
{\color{\colorchanges}
The performance of cMoE is checked at each iteration by computing $\text{CV}({\Omega})$; it is expected that the estimated LOOCV $\text{CV}({\Omega})$ decreases until a minimum, before beginning to increase continuously. This minimum is found by checking whether for $n_{it}$ iterations, the $\text{CV}({\Omega})$ kept only increasing. If so, the cMoE of $n_{it}$ iterations back is considered to have a global minimum error and is selected as optimized model, and the algorithm is terminated. The cMoE algorithm should terminate after $n_{it}$ iterations if the error continually increases, counting from the iteration where $\text{CV}({\Omega})$ reaching its minimum. In the experiments, a value of $n_{\text{it}}=6$ was considered.
}

{\color{\colorchanges}
\subsection{Evaluation of Expert Process Knowledge Integration}
\label{sss.integration}
In the cMoE model, each gate $g_c$ should reflect the knowledge of each context represented by $\pi_c$, and in the prediction stage, the gates are responsible for automatically determining which context the process is running in and switching to the appropriate expert model.
In fact, the $g_c$ is a probability counterpart of the possibility distribution $\pi_c$. The weakest consistency principle \cite{Zadeh1978} leads to a sufficient condition for checking the consistency between the gate's probability function $g_c$ and the possibility distribution $\pi_c$. It states that a probable occurrence must also be possible to some extent, at least to the same degree. The following inequality can formally express this:
\begin{equation}
g_{ci}<\pi_{ci},\, \forall \, i \in \Psi
\end{equation}
The possibility distribution is a upper bound for the probability distribution \cite{DUBOIS2014}. Each context possibility distribution $\pi_c$ reflects the knowledge expert's uncertainty in a quasi-qualitative manner that is less restrictive than probability $g_c$. This is visually represented in Fig. \ref{fig.cMoLE_context_learning},
\begin{figure}[!t]
   \begin{center}
   \includegraphics[width=1\columnwidth]{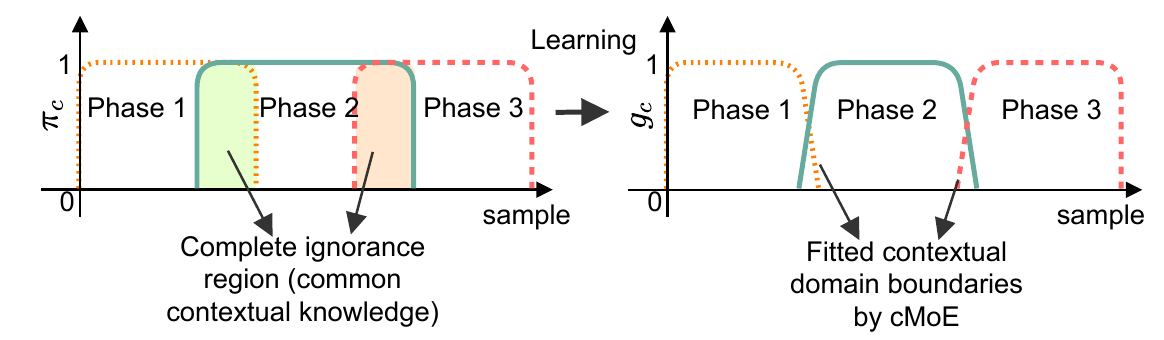}
   \caption{Representation of contextual information with uncertainty (left), and fitted contextual information with cMoE (right).}
 \label{fig.cMoLE_context_learning}
   \vspace{-1.6em}
   \end{center}
 \end{figure}\relax
for a hypothetical three phases process. The left picture shows the contextual information provided by the analyst, with a complete ignorance region (common contextual knowledge). The right picture shows the fitted contextual information with well-defined boundaries, represented as the gate’s output. In the specific case where context information follows the complete ignorance possibility distributions ($\pi_{ci}=1, \text{ for }  c=1,\ldots,C, \text{ and } i=1,\ldots, N$), the cMoE reduces to MoE.

To access if the assignment of context $c$ is correct, the following consistency index of context $c$ ($C_{I,c}$) is defined:
\begin{align}
\label{equ.cic}
C_{I,c}  = \frac{1}{N}\sum_{i=1}^N{I}(g_{ci},\pi_{ci})
\end{align}
where $I(a,b)=1$ iff $a\leq b$, and $0$ otherwise. This consistency index measures the accuracy of the gates' agreement of context $c$ with respect to the consistency principle. To measure the consistency of all contexts in representing the expert knowledge, the following geometric mean is employed 
\begin{equation}
\label{equ.ci}
C_{I}= \left( \prod_{c=1}^{C}C_{I,c} \right)^{1/C}
\end{equation}
If $C_{I}=1$ indicates complete agreement, $C_{I}=0$ indicates complete disagreement, indicating an inability of cMoE to incorporate the expert knowledge into its structure.
In such cases, the uncertainty in the expert knowledge from the possibility distributions can also be re-tuned, for example in $\alpha$-Certain distribution, the certainty $\alpha$ parameter can be tuned from automatic methods. In this case, the $\alpha$ should be chosen using the minimal specificity principle, i.e. search for the most informative distribution (lowest $\alpha$), while keeping a desired consistency index. This can be stated as
\begin{align}
\alpha &= \inf \left\{\alpha^* \in [0,1]: C_I(\alpha^*) < \epsilon \right\}.
\label{equ.ciprinciple}
\end{align}
{\color{black} where $0\leq\epsilon \leq1$ is the minimum desired consistency index.}
The same reasoning can be applied to the $\beta$-Trapezoidal possibility distribution.
}

\section{Experimental Results}
\label{s.experimental}
This section presents the experimental results in two industrial case studies. The first case study deals with the estimation H\(_2\)S at the output stream of a sulfur recover unit described in \cite{Fortuna2007}. The second case study predicts the acidity number in a multiphase polymerization reactor.

\paragraph{State of art models} The following models were also implemented along with the experiments for performance comparison purposes. The MoLE with LASSO penalty \cite{Souza2021}, the LASSO regression model, PLS regression model, Gaussian mixture regression model (GMR), decision tree (TREE), and the optimally pruned extreme learning machine model (ELM) \cite{Miche2010}. {\color{black} The MoLE and LASSO source code are based on the MoLE Toolbox available at \cite{Souza2021}. The PLS is based on the author's own implementation. The GMR implementation is based on the Netlab Toolbox available at \cite{Nabney2022}. The TREE is based on the Matlab implementation of the Statistics and Machine Learning Toolbox. The ELM were implemented from the author's source code, available at \cite{ELM}.

{\color{black}\paragraph{Hyper-parameters Selection}The selection of MoLE and LASSO {\color{black} regularization parameters} follows the predicted LOOCV error described in Sec. \ref{sss.experts_selection}; the LASSO parameter is denoted as $\lambda$, while the regularization parameters of MoLE is defined as $\lambda^e_p,\lambda^g_p$, whereas the $e,g$ superscript denotes the expert and gates parameters, respectively, while $p$ refers to the expert/gate number}. The selection of PLS latent variables $N_{lat}$ follows from a 10-fold CV error on the training data. The selection of the hidden neurons $N_{neu}$ in the ELM model follows from the optimization procedure described in \cite{Miche2010}. The number of components $N_c$ in the GMR is set equal to the number of contexts. For the tree, the minimum number of leaf node observations $N_{leaf}$, was set to be $N_{leaf}$ for both experiments.}

\paragraph{Experimental settings} The predictive performance is accessed by the root mean square error (RMSE), the coefficient of determination ($\text{R}^2$), and the maximum absolute error (MAE). The results of the second case study have been accessed by following a leave-one-batch-out procedure. The models were trained from all batches except one (to be used as a test). This procedure was repeated so that all the batches were used in the testing phase. The performance metrics were averaged and then reported as the final values. {\color{\colorchanges} Also, a randomization t-test (from \cite{Voet1994}) was used to compare the cMoE performance (the null hypothesis assumes that the RMSE of cMoE and the method to be compared are equal; i.e.\ equal mean), the $p$-value is then reported, if $p\text{-value}<0.05$ the null hypothesis is rejected.} Along with the training procedure, the data for the training data was auto-scaled, and the testing data were re-scaled according to the training parameters (mean and variance). 

\subsection{SRU Unit}
The sulfur recovery unit (SRU) unit aims to remove pollutants and recover sulfur from acid gas streams. The SRU plant takes two acid gas as input, the first (MEA gas), from gas washing plants, rich in H\(_2\)S, and the gas from sour water stripping plants (SWS gas), rich in H\(_2\)S and NH\(_3\) (ammonia). The acid gases are burned into reactors, where H\(_2\)S is transformed into sulfur after oxidation reaction with air. Gaseous products from the reaction are cooled down, collected, and further processed, leading to the formation of water and sulfur. The remaining gas non-converted to sulfur (less than 5\%) is further processed for a final conversion phase. The final gas stream contains H\(_2\)S and SO\(_2\), and online analyzers measure these quantities. The goal is to use the process data to build a soft sensor to replace the online analyzers when under maintenance. 

Five main variables are collected, $X_1$ is the gas flow in MEA zone, $X_2$ is the air flow in MEA zone to the combustion of MEA gas (set manually by the operators), $X_3$ is the airflow in MEA zone regulated by an automatic control loop according to the output stream gas composition, $X_4$ the airflow in SWS zone (set manually by operators), and $X_5$ the gas flow in SWS zone. The target/output is set here to be the H\(_2\)S at the end-tail; the SO\(_2\) can also be predicted using the same principle presented here. Also, according to \cite{Fortuna2007} the operators are more interested in the models that can predict the H\(_2\)S peaks.

\begin{figure}[!t]
\label{fig.sru_process}
  \centering
  \subfigure[ ]
  {\label{fig.sru_reference_data}\includegraphics[width=.48\columnwidth]{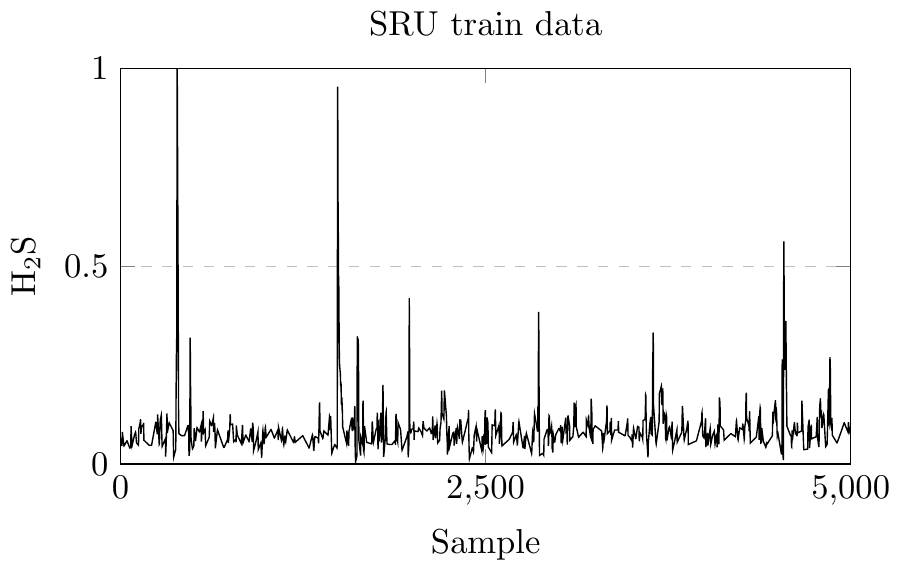}  }
  \subfigure[ ]
  {\label{fig.sru_alpha_ci_loocv}\includegraphics[width=.48\columnwidth]{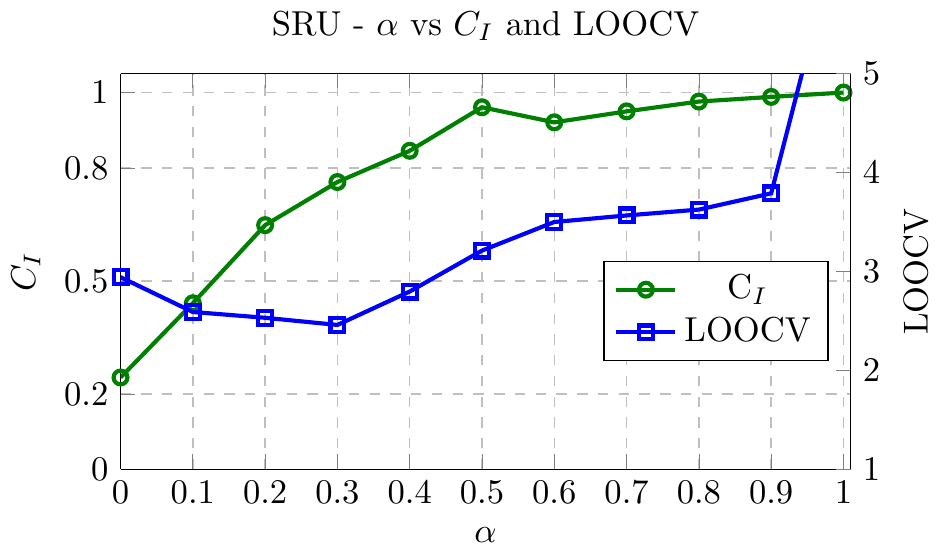} } 
  \subfigure[ ]
  {\label{fig.sru_context_train}\includegraphics[width=.48\columnwidth]{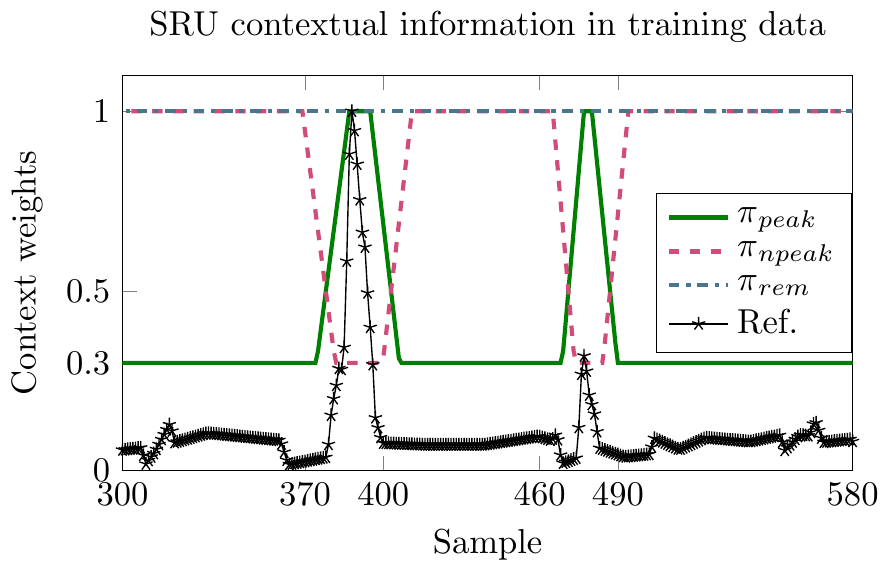} } 
   \subfigure[ ]
  {\label{fig.sru.sru_gates_test}\includegraphics[width=.48\columnwidth]{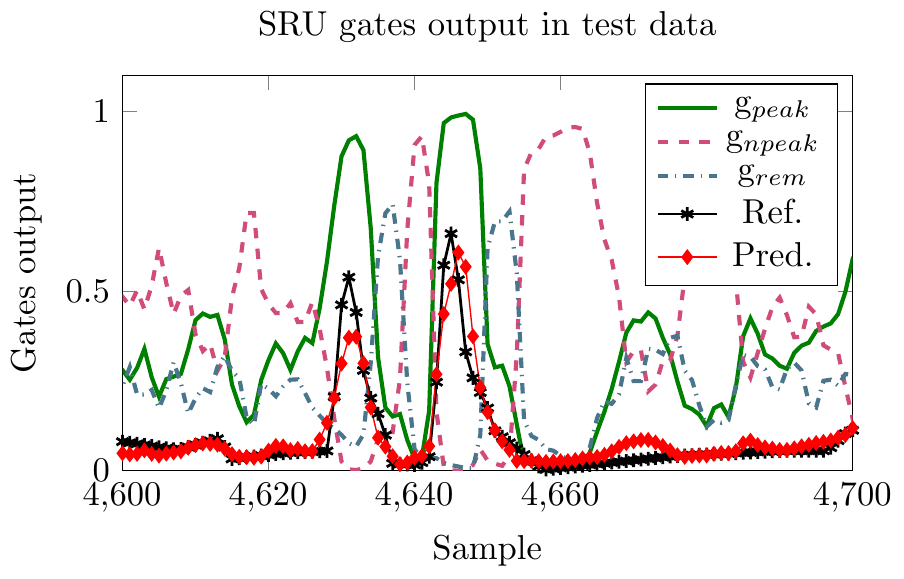} }
  \subfigure[ ]
  {\label{fig.sru.sru_gates_cmole}\includegraphics[width=.48\columnwidth]{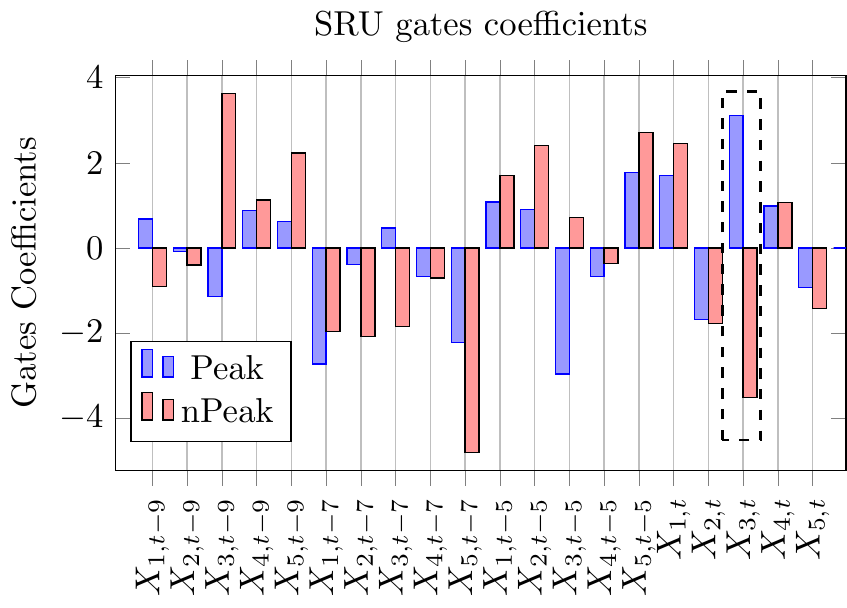} }
  \subfigure[ ]
  {\label{fig.sru.sru_variable_x3}\includegraphics[width=.48\columnwidth]{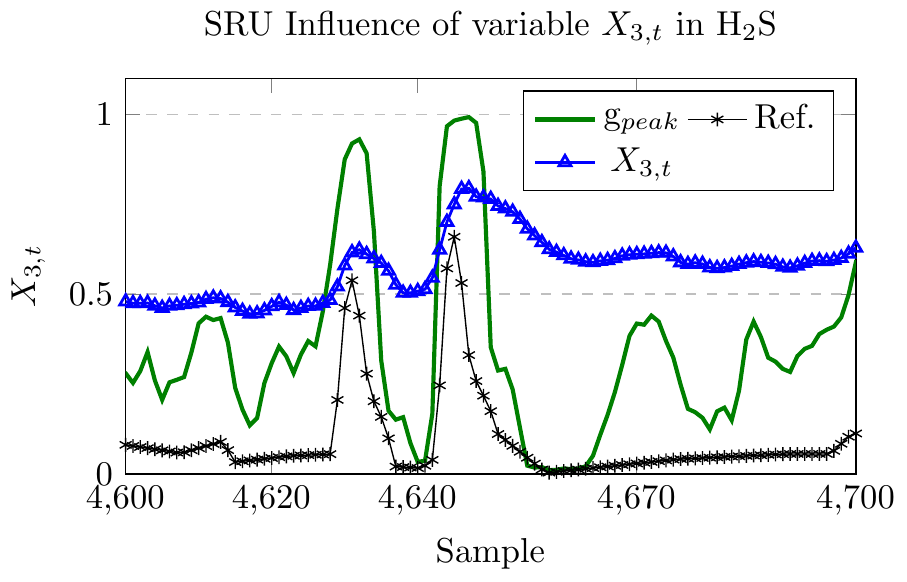} }
  \caption{SRU dataset (a) training data, (b) $\beta$, vs consistency index and LOOCV, (c) contextual information set in training data, and (d) gates output prediction in test data, (e) gates coefficients for peaks and npeaks contexts, (f) effect of variable $X_{3,t}$ in H\(_2\)S output.} 
  \vspace{-1.7em}
\end{figure}

There is a total of $10000$ samples available. The first $5000$ were used for training, and the remaining $5000$ for test; Fig. \ref{fig.sru_reference_data} shows the training data for the SRU dataset, the peaks are clearly visible. Time-lagged features were designed to account for the process dynamics. Then, for the five variables, the time lags of $X_{i,t-d}$, for variables $i=1,\ldots,5$ and delays $d=\{0,5,7,9\}$ were considered, resulting in a total of 20 features. \vspace{-1em}
\relax

To predict the H\(_2\)S, and allow a better understanding of the causes of the peaks, a cMoE model with three contexts was designed. The first, representing the operator context, accounts for the peaks. The second context is designed to represent the non-peaks. The third context represents the remaining process states that are not accounted for the peaks and non-peaks. The cMoE model is represented by
$$\hat{y}=h_{\text{peak}} + h_{\text{npeak}} + h_{\text{rem}}, $$
where ``npeak'' is a typo to non-peaks.
{\color{\colorchanges} Two $\beta$-Trapezoidal possibility distributions were designed for peaks context $\pi_{\text{peak}}$, and non-peaks context $\pi_{\text{npeak}}$, while for the remaining context $\pi_{\text{rem}}$, a complete ignorance distribution is assumed, i.e. $\pi_{\text{rem}}=1, \forall i\in \{1,\ldots,5000\}$ , as there are no information about the context. The peaks were selected manually in the training set, and the peak distribution was designed so that the limits of the trapezoidal function were defined to guarantee that the highest values have $\pi_{\text{peaks}}=1$ peaks. The possibility distribution for the non-peaks was designed to be complementary to the context of the peak, with a lower bound defined by the certainty $\beta$. }
A portion of the expert knowledge feed to cMoE is depicted in Fig. \ref{fig.sru_context_train}. From sample $370-400$ a peak in H\(_2\)S is present (as Ref.). $\pi_{\text{peaks}}$ (see Fig. \ref{fig.sru_context_train} samples $370-400$, and $460-490$). The context of non-peak ($\pi$\(_{npeak}\)) was designed to be complementary to the peaks context. The remaining context ($\pi$\(_{rem}\)=1) for all samples.
{\color{\colorchanges} The uncertainty $\beta$ was chosen using the consistency principle described in Eq. \req{equ.ciprinciple}. The consistency index and the LOOCV, for different values of the uncertainty parameter $\alpha$, are shown in Fig. \ref{fig.sru_alpha_ci_loocv}. The results show that $\beta=0.3$, the selected uncertainty factor, has the lowest LOOCV, with a consistency index of $C_I=0.79$. 
}

Figure \ref{fig.sru.sru_gates_test}, shows the gates output prediction for the peaks (G\(_{peak}\)), non-peaks (G\(_{npeak}\)) and remaining (G\(_{rem}\)) contexts in the test set, for samples $4600-4700$. There are two peaks in this portion of the test data, between samples $4620-4640$, and $4640-4660$. The gates of peaks expert G\(_{peak}\) follow the peak pattern by assigning higher contributions to the peak expert model when the peaks are present. The same behavior is perceived in the non-peaks gates G\(_{npeak}\), which seems to work complementary to the peaks component. The remaining context gates, G\(_{rem}\), seem somewhat oscillating between the patterns; this seems to be related to a constant operation of the system. 
The gates coefficients allow identifying the root causes of change between the peaks and non-peaks. The gates coefficients for the peaks, and non-peaks is shown in Fig. \ref{fig.sru.sru_gates_cmole}. The variable $X_3$ (marked with a rectangle dashed line) has the largest difference between the two gates, indicating this is the main variable that acts on the peaks and non-peaks model switching. 
Figure \ref{fig.sru.sru_variable_x3} shows the variable $X_3$, together with the H\(_2\)S peaks. $X_3$ represented the input airflow to control the end tail H\(_2\)S. Thus H\(_2\)S is a consequence of $X_3$. It seems that the control system is unstable, and any oscillation in $X_3$ causes a large oscillation in the H\(_2\)S. One possible solution to improve the stability of H\(_2\)S and reduce peaks is to improve the control system.
Improvements to the control system can have a positive environmental impact by lowering H\(_2\)S  emissions and/or reducing costs associated with H\(_2\)S post-treatment.

The accuracy of the cMoE was compared with the other state of the art models, and the results are shown in Table \ref{table.sru}.
\begin{table}[!t]
\caption{H\(_2\)S prediction accuracy for all compared models}
\label{table.sru}
\setlength{\tabcolsep}{5.3 pt}
\begin{tabular}{l|ccccccc}
\toprule
 \multicolumn{8}{c}{\textbf{H\(_2\)S}}\\
 \hline
 & cMoE & MoLE & LASSO & PLS & GMR & TREE & ELM \\
 $\text{R}^2$ & \bfn{0.732} & 0.583 & 0.085 & 0.541 & 0.519 & 0.001 & 0.002 \\
 $\text{RMSE}$ &  \bfn{0.026} & 0.035 & 0.053 & 0.035 & 0.032 & 0.087 & 0.060  \\
 $\text{MAE}$ & \bfn{0.340} & 0.484 & 0.630 & 0.519 & 0.401 & 0.603 & 0.662 \\
 p-value & 1.000 & 0.001 & 0.001 & 0.001 & 0.001 & 0.001 & 0.001 \\
\bottomrule
\end{tabular}
\vspace{-1.5em}
\end{table}
The results show that the cMoE outperforms all the other models with statistical significance. Results confirm that constraining the model to represent the system’s expected behavior positively impacts the prediction performance. 
{\color{black}Table \ref{table.resultsParH2S}
\begin{table*}[!t]
\centering
\caption{Hyper-parameters of the fitted H\(_2\)S models}
\label{table.resultsParH2S}
\begin{tabular}{ccccccc}
\toprule
 \multicolumn{7}{c}{\textbf{H\(_2\)S}}\\
 \hline
cMoE & MoLE & LASSO & PLS & GMR & TREE & ELM \\
$\lambda^e_{\text{peak}}=0.00$,$\lambda^g_{\text{peak}}=0.00$  & $\lambda^e_{\text{peak}}=1.00$,$\lambda^g_{\text{peak}}=1.049$ & \multirow{3}{*}{$\lambda=1.00$} & \multirow{3}{*}{$N_{lat}=16$} & \multirow{3}{*}{$N_{c}=3$} & \multirow{3}{*}{$N_{leaf}=1$} & \multirow{3}{*}{$N_{neu}=160$} \\
$\lambda^e_{\text{npeak}}=1.00$,$\lambda^g_{\text{npeak}}=3.34$  & $\lambda^e_{\text{npeak}}=12.61$,$\lambda^e_{\text{npeak}}=1.31$ & & & & & \\
    $\lambda^e_{\text{rem}}=1.00$,$\lambda^g_{\text{rem}}=1.00$  & $\lambda^e_{\text{rem}}=12.83$,$\lambda^e_{\text{rem}}=0.00$ & & & & & \\
\bottomrule
\end{tabular}
\vspace{-1.5em}
\end{table*}\relax
shows the parameters obtained for each model in the  H\(_2\)S experiment.
}
\vspace{-1em}

\subsection{Polymerization}
This case study refers to a polymerization batch process for resin production. The material is loaded into the reactor, which then undergoes the five process phases: heating, pressure, reaction, vacuum, and cooling; most of the phase changes are triggered manually by the operators. The phase change is determined by the quality values of the resin, namely the resin acidity number and the resin viscosity. While a physical sensor measures the viscosity, the acidity number is measured three times, one at the vacuum phase and two at the reaction phase. The objective here is to build a soft sensor to measure the acidity number online and better understand the variables that affect the acidity in that two phases. 

For this process, there are data for $33$ batches in the specification, with a total of $17$ variables measured along the process; they are described in Table \ref{tab.variables_dsm}. As there are three acidity measurements for each batch, a total of $99$ samples is available. The process variables are synchronized with the acidity number by removing the samples that do not have the corresponding acidity number values. 
\begin{table}[!t]
\caption{Variables of the polymerization unit.}
\label{tab.variables_dsm}
{\scriptsize
\begin{center}
\begin{tabular}{clcl}
\toprule
     Variable & Description & Variable & Description\\ 
\hline 
     $\text{X}_1$  &  Reactor temperature; & $\text{X}_{10}$  &  Reactor pressure 2;   \\ 
     $\text{X}_2$  &  Temperature oil return; &  $\text{X}_{11}$  &  Pressure reflux;   \\ 
     $\text{X}_3$  &  Temperature gas return; & $\text{X}_{12}$  &  Vacuum presure;   \\ 
     $\text{X}_4$  &  Temperature reflux pump; & $\text{X}_{13}$  &  Flow reflux;   \\ 
     $\text{X}_5$  &  Condensator cooling temperature; & $\text{X}_{14}$  &  Flow oil;   \\ 
     $\text{X}_6$  &  Column temperature; & $\text{X}_{15}$  &  Lever water;   \\ 
     $\text{X}_7$  &  Reactor temperature mixture; & $\text{X}_{16}$  &  Level solvent;  \\ 
     $\text{X}_8$  &  Temperature thermal oil ; &$\text{X}_{17}$  &  Viscosity;   \\ 
     $\text{X}_9$  &  Reactor pressure 1; & $\text{Y}$  &  Acidity number.  \\ 
\bottomrule
\end{tabular}
\vspace{-1.5em}
\end{center} }
\end{table}\relax

Then, a cMoE model with two contexts was designed to predict acidity and understand the variables that mostly affect the acidity number. The first context represents the reaction phase, and the second to the vacuum phase. The cMoE model for acidity prediction is represented by
$$\hat{y}=h_{\text{reaction}} + h_{\text{vacuum}} $$

\begin{figure}[!t]
\centering
\subfigure[ ]
{\label{fig.dsm.dsm_context_train}\includegraphics[width=.48\columnwidth]{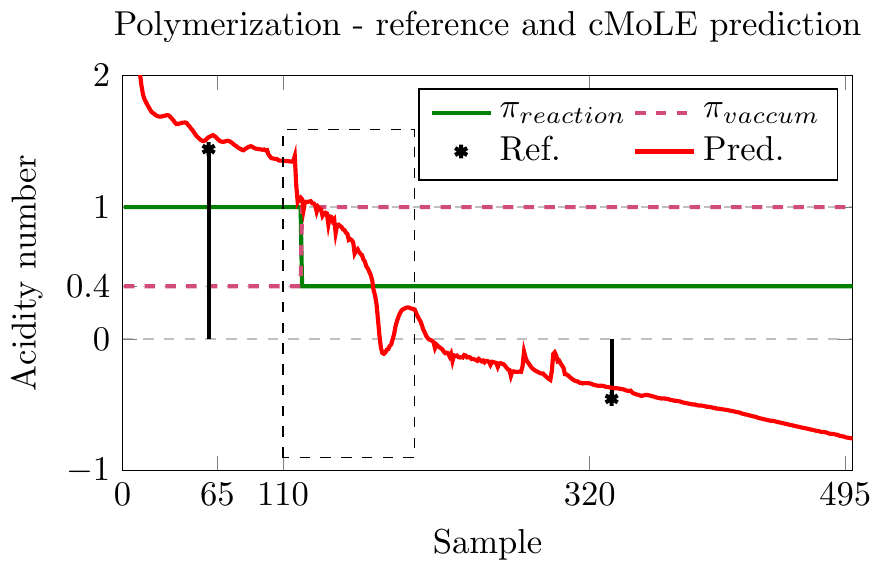}  }
\subfigure[ ]
{\label{fig.dsm.dsm_cmole_experts_gates}\includegraphics[width=.48\columnwidth]{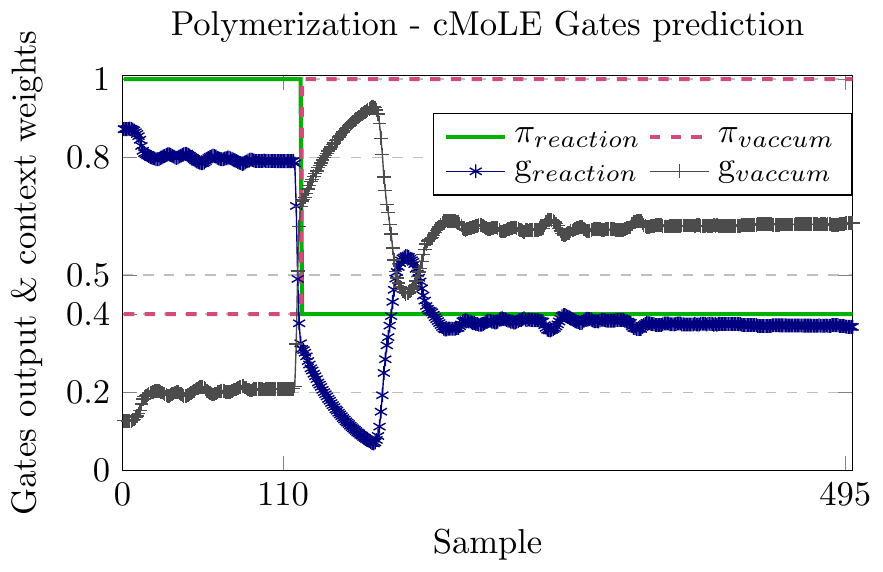} } 
\subfigure[ ]
{\label{fig.dsm.dsm_alpha_loocv_ci}\includegraphics[width=.48\columnwidth]{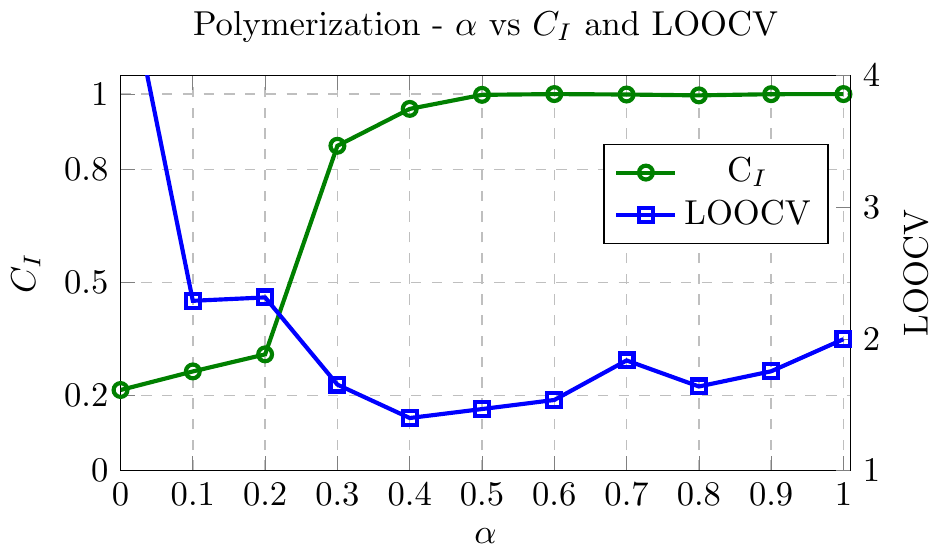} }
\subfigure[ ]
{\label{fig.dsm.dsm_cmole_experts_coeff}\includegraphics[width=.48\columnwidth]{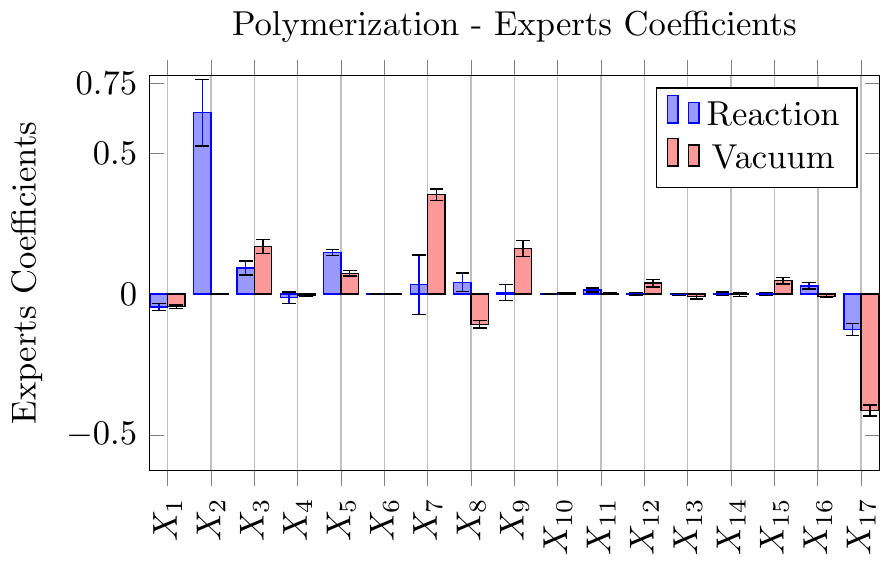} } 
\subfigure[ ]
{\label{fig.dsm.dsm_cmole_gates_coefficients}\includegraphics[width=.48\columnwidth]{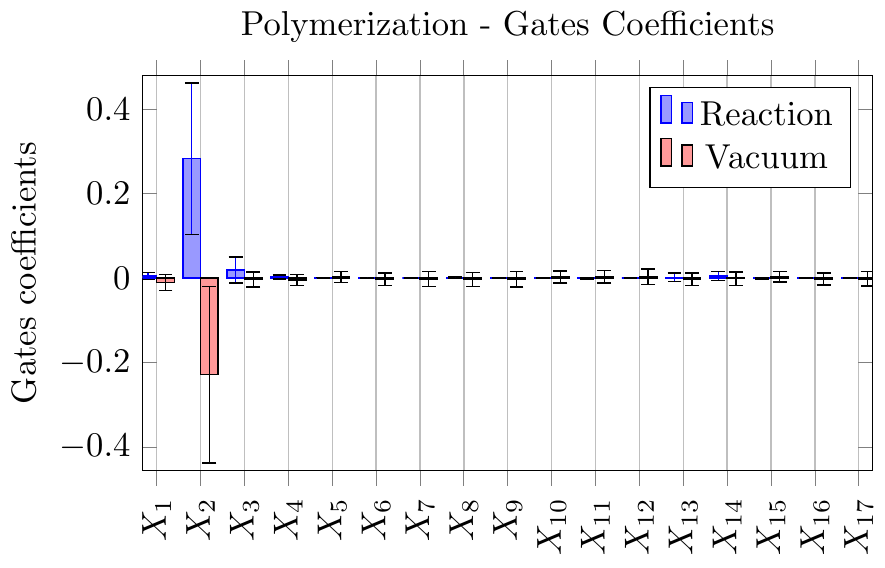} }
\subfigure[ ]
{\label{fig.dsm.dsm_cmole_variable_x2}\includegraphics[width=.48\columnwidth]{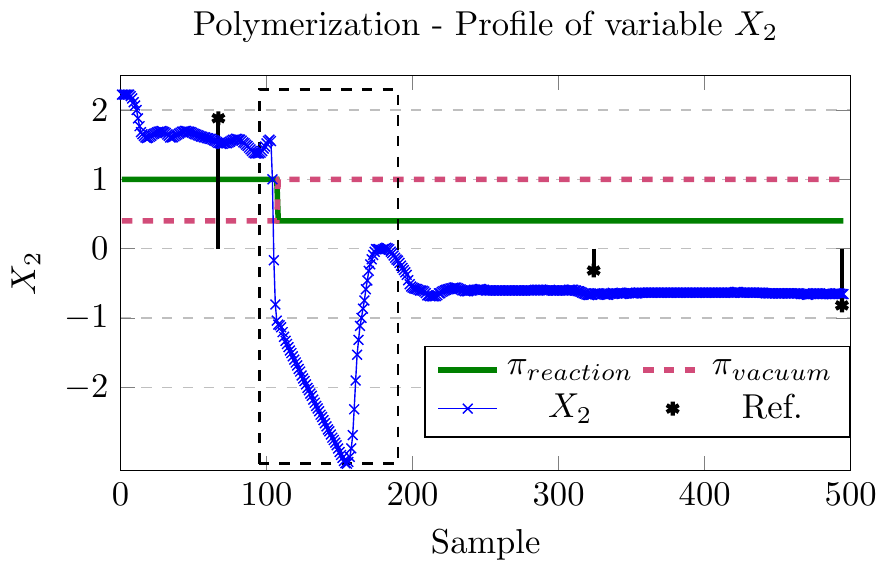} }
\caption{Polymerization dataset, a) context for reaction and vacuum phases, together with acidity and prediction by cMoE, b) cMoE gates output, c) $\alpha$, vs consistency index and LOOCV, d) reaction and vacuum experts coefficients, e) reaction and vacuum gates coefficients, f) variable $X_2$ (temperature oil return) } 
\label{fig.dsm_coefficients}
\vspace{-1.2em}
\end{figure}

The process knowledge is available as the phase changes from the operators, between the reaction and vacuum phases. {\color{\colorchanges} In this case, the $\alpha$-Certain distribution was designed to represent the operator's context of the phases.} This is depicted in Fig. \ref{fig.dsm.dsm_context_train} for a single batch. There, the two contexts {\color{\colorchanges}($\pi_{\text{reaction}}$ and $\pi_{\text{vacuum}}$)} taken by operators indicate the region of samples belonging to each phase; the change between phases occurs at sample $110$. The acidity number is also indicated as `Ref.', measured at samples $65$, $320$, and $495$. {\color{\colorchanges} The uncertainty $\alpha$ was chosen using the consistency principle described in Eq. \req{equ.ciprinciple}. The consistency index and the LOOCV, for different values of the uncertainty parameter $\alpha$, are shown in Fig. \ref{fig.dsm.dsm_alpha_loocv_ci}. The results show that $\alpha=0.4$ has the lowest LOOCV, with a consistency index of $C_I=0.99$. It is worth noting that the LOOCV error is significantly higher for $\alpha=0$ (no uncertainty) when compared to higher uncertainty $\alpha>0$, indicating that uncertainty plays a significant role in representing process expert knowledge.}
\relax

The predictive performance from the leave-one-batch-out procedure, for the all models compared, is shown in Table \ref{table.dsm_results}, in terms of $\text{R}^2$, RMSE and MAE.
\begin{table}[!t]
\caption{Acidity prediction accuracy for all compared models}
\setlength{\tabcolsep}{5.3 pt}
\begin{tabular}{l|ccccccc}
\toprule
 \multicolumn{8}{c}{\textbf{Acidity}}\\
 \hline
  & cMoE & MoLE & LASSO & PLS & GMR & TREE & ELM \\$\text{R}^2$ & \textbf{0.996} & 0.995 & 0.995 & 0.994 & 0.996 & 0.994 & 0.812 \\$\text{RMSE}$ & \textbf{0.092} & 0.101 & 0.101 & 0.109 & 0.096 & 0.120 & 0.500 \\$\text{MAE}$ & \textbf{0.134} & 0.155 & 0.155 & 0.166 & 0.145 & 0.174 & 0.768 \\$p$-value & 1.000 & 0.342 & 0.350 & 0.010 & 0.669 & 0.028 & 0.001 \\
\bottomrule
\end{tabular}
\vspace{-1.5em}
\label{table.dsm_results}
\end{table}
{\color{black}Table \ref{table.dsm_results}
\begin{table*}[!t]
\centering
\caption{Hyper-parameters of the fitted Acidity models}
\label{table.dsm_results}
\begin{tabular}{ccccccc}
\toprule
 \multicolumn{7}{c}{\textbf{Acidity}}\\
 \hline
cMoE & MoLE & LASSO & PLS & GMR & TREE & ELM \\
$\lambda^e_{\text{reaction}}=1.00$,$\lambda^g_{\text{reaction}}=1.15$  & $\lambda^e_{\text{reaction}}=16.90$,$\lambda^g_{\text{reaction}}=0$ & \multirow{2}{*}{$\lambda=31.90$} & \multirow{2}{*}{$N_{lat}=5$} & \multirow{3}{*}{$N_{c}=2$} & \multirow{2}{*}{$N_{leaf}=1$} & \multirow{2}{*}{$N_{neu}=106$} \\
$\lambda^e_{\text{vacuum}}=21.92$,$\lambda^g_{\text{vacuum}}=2.36$  & $\lambda^e_{\text{vacuum}}=16.90$,$\lambda^e_{\text{vacuum}}=0$ & & & & & \\
\bottomrule
\end{tabular}
\vspace{-1.5em}
\end{table*}\relax
shows the parameters obtained for fitted models for the first fold of the leave-one-batch-out procedure.
}
The cMoE is statistically different at a $0.05$ significance level than PLS, TREE, and ELM and has superior performance to all other models. Also, when inspecting the gate’s output provided by cMoE, the cMoE model significantly retained the representation of the initial contextual information and detected the change between phases as shown by the gate’s output in Figs. \ref{fig.dsm.dsm_cmole_experts_gates}.

The cMoE coefficients for the reaction, and vacuum experts are shown in Fig. \ref{fig.dsm.dsm_cmole_experts_coeff}. The reaction expert has a significant representation of the reaction's phase of the polymerization unit. The main variables which were important for the reaction expert are the oil temperature ($\text{X}_2$, $\text{X}_3$), reactor mixture temperature ($\text{X}_7$), condenser temperature ($\text{X}_5$) and the liquid viscosity ($\text{X}_{17}$). The vacuum expert represents a more significant portion of the vacuum of the polymerization unit. The most significant variable is the viscosity ($\text{X}_{17}$), as the viscosity is physically a function of pressure. It is also important on the gas return temperature ($\text{X}_3$), reactor temperature, and pressure ($\text{X}_7$, $\text{X}_9$). These variables are physically related to the equation of states of the gaseous product within the reactor. 

To better understand the phases transition Fig. \ref{fig.dsm.dsm_cmole_gates_coefficients} shows the gates coefficients for the reaction and vacuum contexts. From there, variable $X_2$, the temperature of oil return, mostly affects the transition to phases. The variable $X_2$ is plotted in Fig. \ref{fig.dsm.dsm_cmole_variable_x2}. From there, it is possible to check that there is an intermediate step in which oil temperature drops up to a minimum before the operator starts the vacuum phase. The process status again, reaching normalization in sample $200$. Also, the oil temperature has two different regimes in the reaction and vacuum phases. 
\vspace{-.13em}
\section{\color{\colorchanges}Discussion}
{\color{\colorchanges}
The proposed cMoE model uses possibility distributions to represent contextual information provided by process operators and to integrate this expert knowledge into the cMoE model via the data using the learning procedure. In addition, to assess how well expert knowledge was integrated into the cMoE model, a consistency index was defined in Eq.\ \req{equ.ci}. The first case study, a continuous process, was broken down into three contexts, with the peaks and non-peaks contexts being the most relevant to operators. As a result, important information and insights on the status of the control system were obtained by inspecting the main variables causing the transition between the peaks and non-peaks contexts. In the second case study,  a multiphase batch process, the information on phase transitions was the knowledge to be integrated into the model. As a result, more insights into phase transitions and the impact of each variable on each phase were realized. 

It is worth noting that the linear models in cMoE are sufficient to integrate the expert knowledge in both case studies; also worth mentioning that the uncertainty parameters in both cases studies were refined from the data, together with an analysis of the consistency index. In cases where the consistency index performs poorly, or one wants to employ a more informative distribution (i.e., lower values of uncertainty), non-linear models for gates and experts may be employed as an alternative to linear models so that the consistency index's performance is improved. It would be expected that non-linear modeling must capture the non-linear behavior of the data that must be relevant in integrating expert knowledge. Furthermore, the $\ell_1$ penalty is not required for the linear solution of cMoE; other penalties, such as $\ell_2$ or $\ell_{1,2}$, can also be employed. In the case of data collinearity, the solution can be obtained by applying the PLS model to experts and gates, as demonstrated in \cite{Souza2014a}.

Compared to other models that are interpretable by nature, such as the Lasso, PLS, DT, and GMR, they lack mechanisms to integrate expert knowledge. Of course, meaningful relations and rules can be extracted from these models, but this is still driven by the data, and if no context is provided, the extraction of relevant information for more complex relationships, as in the two case studies presented here, is not possible. The contextual framework presented here is flexible enough to allow its implementation in many models, including the GMR model.

}

\section{Conclusions}
\label{s.conclusions}
In conclusion, this paper proposes the contextual mixture of experts model, a data-driven model devised to incorporate operator domain knowledge into its structure. The proposed approach has been shown to increase predictive performance while achieving a direct interpretation of process variable contribution in each regime of the process. This approach was evaluated on two different problems, demonstrating better statistical performance than conventional machine learning models that do not rely on contextual information. The proposed method has strong potential as a stable and explainable framework to include contextual information in data-driven modeling. This is important to help the transition to Industry 5.0 by increasing the human-machine synergy in the process industry.  
{\color{\colorchanges}
Future research could concentrate on non-linear functions for experts and gates learning to improve predictive performance, as well as explainable methods for interpretability.}

\bibliographystyle{IEEEtran}
\bibliography{\jobname}

\WarningFilter{latex}{Overwriting file}
\WarningFilter{latex}{Tab has been converted to Blank Space}

\ActivateWarningFilters[latex]

\end{document}